# A Robust Matching Pursuit Algorithm Using Information Theoretic Learning

Miaohua Zhang[1], Yongsheng Gao[1], Changming Sun[2], and Michael Blumenstein[3]

[1]School of Engineering and Built Environment, Griffith University, Australia
[2]CSIRO Data61, NSW, Australia
[3]Faculty of Engineering and IT, University of Technology Sydney, Australia

# Abstract

Current orthogonal matching pursuit (OMP) algorithms calculate the correlation between two vectors using the inner product operation and minimize the mean square error, which are both suboptimal when there are non-Gaussian noises or outliers in the observation data. To overcome these problems, a new OMP algorithm is developed based on the information theoretic learning (ITL), which is built on the following new techniques: (1) an ITL-based correlation (ITL-Correlation) is developed as a new similarity measure which can better exploit higher-order statistics of the data, and is robust against many different types of noise and outliers in a sparse representation framework; (2) a non-second order statistic measurement and minimization method is developed to improve the robustness of OMP by overcoming the limitation of Gaussianity inherent in cost function based on second-order moments. The experimental results on both simulated and real-world data consistently demonstrate the superiority of the proposed OMP algorithm in data recovery, image reconstruction, and classification.

# 1. Introduction

Sparse representation (SR) aims to find a subset of atoms in an over-complete dictionary to represent the query sample by giving larger weights to the selected atoms and smaller (or nearly zero) weights to the remaining atoms [1-3]. Because of this property, SR is often used as a useful feature selection tool for object recognition [4-6], signal recovery [7], visual tracking [8], image retrieval [9], and others. Matching pursuit family is one group of sparse representation methods that select a subset of features (atoms) from the training data (dictionary) for learning in a greedy fashion. One representative example of this family is the matching pursuit algorithm [10] which provides an efficient solution to solve the SR problem by iteratively selecting the highest-correlated basis vectors (atoms) in a dictionary. An important advancement on matching pursuit is orthogonal matching pursuit (OMP) [11], in which all the coefficients are updated after every step of iteration by an orthogonal projection to exclude the information of selected atoms, which produces better results than the standard matching pursuit, but it is more computationally expensive. To improve the efficiency and performance of OMP, many variants with different feature selection or residual minimization strategies have been proposed in the literature. For example, different from OMP that selects one atom in each iteration, Wang *et al*. [12] select multiple atoms to improve the efficiency of identifying atoms and proposed a generalized OMP (GOMP) method. Owing to the multiple-atom selection strategy, the GOMP algorithm greatly reduces the number of iterations required for convergence. Unlike GOMP, the compressive sampling matching pursuit (CoSaMP) [13] reduces the running time by introducing a pruning step after the estimation step, which can cuts down the number of the selected atoms and speeds up the subsequent least-squares computation, and also maintain the sparse approximation at the same time.

Tree search and neural network based matching pursuit is another important group which mainly focuses on learning deep features from multiple paths [14-16] or from single hidden neural network [17]. In [14], the authors proposed a multipath hierarchical matching pursuit to learn features by capturing multiple aspects of discriminative structures of the data in a deep path architecture. Algorithms in [15] and [16] are tree search based methods which use different deep tree search strategies during feature selection and estimation procedures to improve the sparse approximation. In [17], the authors proposed a learned OMP algorithm for fast sparse approximation by learning a single hidden neural network for subspace clustering.

However, these existing MP algorithms estimate the sparse vectors by minimizing a mean square error (MSE) based loss function, whose performances can be significantly deteriorated when the data are corrupted by outliers, which is inevitable in real-world applications. MSE-based loss function assigns same weights to all measures without any discriminative constraints on ever severely or slightly corrupted ones when minimizing the residual. However, outliers are typically far away from the centre of the normal data, thus such an equal weight assignment will result in an incorrect sparse solution due to the significant influence of those corrupted measures. Moreover, the MSE-based loss function assumes that noise/outliers follows the Gaussian distribution, which is not only sensitive to non-Gaussian noise but also to outliers [18-20]. As a consequence, these existing methods fails to approximate the sparse coding if the assumption does not hold [21, 22]. To solve this problem, some robust functions have been proposed which include arbitrary-order based functions and information theoretic learning (ITL)-based methods [21-26]. In [23], although the author proposed a robust function in the $l_p$-spaces to improve its robustness against noise and outliers, their model is time consuming when processing high-dimensional image data. The correntropy matching pursuit (CMP) algorithm [24] overcomes the above drawbacks by taking advantage of the correntropy in reducing the effect from outliers. However, CMP still suffers from noise and outlier corruption because it minimizes the reconstruction error by a second-order kernel loss function. Furthermore, the existing matching pursuit algorithms, including CMP, calculates the correlation (similarity) between two vectors by the inner product operation. However, the inner product is not a noise/outlier resistant function since it is defined in the $l_2$-space, which implies that using an inner product operation for feature selection in matching pursuit algorithms will be sensitive to outliers [23].

To solve problems mentioned above, we proposed a new matching pursuit method that is robust to many types of non-Gaussian noise and outlier in the observation data in this paper. An early conference version of this research is reported in [27]. In this paper, we present our complete work with new and improved formulations, extended experimental investigation and analysis. Our work contributes to robust OMP methods from the following three aspects:

- We developed a new ITL-Correlation based on the information theoretic learning to measure the similarity and difference between two samples in the presence of noise and/or outliers. Different from the inner product or cross correlation used in existing matching pursuit algorithms, which is only correct under Gaussian conditions, the ITL-based correlation function is robust against heavy-tailed impulsive noise that is commonly associated with large-amplitude outliers.
- Traditional MP algorithms measure the representation error by MSE, which highly relies on that the noise obey a Gaussian distribution, and thus these algorithms fail when noise/outliers follow a non-Gaussian distribution. To overcome this problem, in this paper, we proposed a non-second order kernel (NOK) statistic measurement. Different from the second order correntropy used in CMP [24], the proposed NOK-loss gives more flexibility in controlling the reconstruction error and thus obtains much better results for the challenging occlusion problem in object reconstruction and recognition.

- A new classifier based on the NOK-loss is developed to minimize the effect from outliers and non-Gaussian noise for robust classification.

The remainder of this paper is organized as follows: The related works are introduced in Section 2. The proposed algorithm, including the definition of ITL-Correlation and NOK measurement, is introduced in Section 3. Section 4 gives experimental results of our algorithm compared with benchmark methods. Conclusions are drawn in Section 5.

# 2. Related Works

Sparse representation [2] represents a query sample $\boldsymbol{b} \in R^m$ by a linear combination of a subset of samples chosen from an overcomplete dictionary $\boldsymbol{A} \in R^{m \times N}$, which is typically represented by

$$\min_{\boldsymbol{x}} \|\boldsymbol{x}\|_0 \quad \text{subject to} \quad \boldsymbol{b} = \boldsymbol{A}\boldsymbol{x}, \tag{1}$$

where $\|\cdot\|_0$ denotes the $l_0$-norm, counting the nonzero entries in a vector, and $\boldsymbol{x} \in R^N$ is a sparse coefficient having $L$ non-zero entries ($L$-sparse). In the following, we will introduce how does the OMP algorithm solve the above problem. Besides, a new loss function based on the information theoretic learning is also introduced to improve the robustness of OMP against noise/outliers.

## 2.1. Orthogonal Matching Pursuit

The OMP [11] algorithm sequentially chooses dictionary atoms that have the highest correlation with the representation residual in a greedy fashion. The two most important procedures in this algorithm are the correlation calculation between the current residual and the dictionary atoms, and the least squares problem solving component (estimation).

Denoted by $\boldsymbol{a}_i$ the $i$th atom in a dictionary $\boldsymbol{A}$, $S_0 = \Phi$ the initial support set, $<\cdot,\cdot>$ the inner product operation, and $\boldsymbol{r}_{k-1}$ the approximate error at the $(k-1)$th iteration. The correlation between the current residual and dictionary atom at the $k$th iteration can be calculated by

$$i_0 = \arg\max_{\boldsymbol{a}_i \in \boldsymbol{A}} |\langle \boldsymbol{r}_{k-1}, \boldsymbol{a}_i \rangle|, \qquad 1 \le i \le N, \qquad k = 1,2,\cdots. \tag{2}$$

The initial residual is set as $\boldsymbol{b}$, i.e., $\boldsymbol{r}_0 = \boldsymbol{b}$. The solution support is updated by $S_k = S_{k-1} \cup \{i_0\}$. The provisional solution is obtained by minimizing

$$\boldsymbol{x}_{S_k} = \arg\min_{\boldsymbol{x}} \left\| \boldsymbol{A}_{S_k}\boldsymbol{x} - \boldsymbol{b} \right\|_2^2, \tag{3}$$

which is a least-squares (LSs) problem and has a closed-form solution. The solution $\boldsymbol{x}_{S_k}$ is given by $\left(\boldsymbol{A}_{S_k}{}^T \boldsymbol{A}_{S_k}\right)^{-1} \boldsymbol{A}_{S_k}{}^T \boldsymbol{b}$. The algorithm then updates the residual by

$$\boldsymbol{r}_k = \boldsymbol{r}_{k-1} - \boldsymbol{A}_{S_k}\boldsymbol{x}_{S_k}. \tag{4}$$

The algorithm terminates if the algorithm reaches the required sparsity level, or when the norm of the residual is less than a preset threshold.

## 2.2. Correntropy Measurement

Let $\phi(\boldsymbol{v}_1)$ be a nonlinear mapping function which transforms $\boldsymbol{v}_1$ from the original space to the Hilbert space, satisfying $\langle \phi(\boldsymbol{v}_1), \phi(\boldsymbol{v}_2) \rangle_H = k(\boldsymbol{v}_1, \boldsymbol{v}_2)$ [25, 26], the standard correntropy loss (c-loss) is defined by

$$\begin{aligned} J_{\text{c-loss}}(\boldsymbol{v_1}, \boldsymbol{v_2}) &= \frac{1}{2}\mathrm{E}[\|\phi(\boldsymbol{v}_1) - \phi(\boldsymbol{v}_2)\|_H^2] \\ &= \mathrm{E}[1 - k_\sigma(\boldsymbol{v}_1 - \boldsymbol{v}_2)] \\ &= 1 - V(\boldsymbol{v}_1 - \boldsymbol{v}_2), \end{aligned} \tag{5}$$

where $k_\sigma(\cdot)$ and $\mathrm{E}(\cdot)$ are the Gaussian kernel function and mathematical expectation operator. Chen et al. proposed a kernel mean $p$-power error (KMPE) [25] with the generalized ability to handle arbitrary order errors for signal processing and image processing, the KMPE is given by

$$\begin{aligned} J_{\mathrm{KMPE}}(\boldsymbol{v}_1, \boldsymbol{v}_2) &= 2^{-\frac{p}{2}}\mathrm{E}[\ \|\phi(\boldsymbol{v}_1) - \phi(\boldsymbol{v}_2)\|_H^p] \\ &= 2^{-\frac{p}{2}}\mathrm{E}[\|\phi(\boldsymbol{v}_1) - \phi(\boldsymbol{v}_2)\|_H^2]^{p/2} \\ &= 2^{-\frac{p}{2}}\mathrm{E}\left[\left(2k_\sigma(0) - 2V(\boldsymbol{v}_1, \boldsymbol{v}_2)\right)\right]^{p/2} \\ &= \mathrm{E}\left[\left(1 - k_\sigma(\boldsymbol{v}_1 - \boldsymbol{v}_2)\right)^{p/2}\right], \end{aligned} \tag{6}$$

where $p > 0$ controls the representation error and can be adjusted to a suitable value to improve the representation ability and performance. Obviously, the KMPE becomes c-loss function when $p$ is 2.

# 3. Proposed Robust Matching Pursuit Algorithm

## 3.1. Information Theoretic Learning-Based Correlation (ITL-Correlation)

In conventional greedy algorithms, the correlation between two vectors $\boldsymbol{b} = [b_1, \cdots, b_m]^T \in R^m$ and $\boldsymbol{a} = [a_1, \cdots, a_m]^T \in R^m$ are defined as [28, 29]

$$\langle \boldsymbol{b}, \boldsymbol{a} \rangle = b_1 a_1 + \cdots + b_m a_m. \tag{7}$$

The angle between $\boldsymbol{b}$ and $\boldsymbol{a}$ is defined by

$$\theta = \cos^{-1}\left(\frac{\langle \boldsymbol{b}, \boldsymbol{a} \rangle}{\|\boldsymbol{b}\| \|\boldsymbol{a}\|}\right). \tag{8}$$

From the Cauchy-Schwarz inequality, we have

$$\langle \boldsymbol{b}, \boldsymbol{a} \rangle^2 \le \langle \boldsymbol{b}, \boldsymbol{b} \rangle \langle \boldsymbol{a}, \boldsymbol{a} \rangle \Rightarrow \frac{\langle \boldsymbol{b}, \boldsymbol{a} \rangle^2}{\langle \boldsymbol{b}, \boldsymbol{b} \rangle \langle \boldsymbol{a}, \boldsymbol{a} \rangle} \le 1 \Rightarrow \frac{|\langle \boldsymbol{b}, \boldsymbol{a} \rangle|}{\|\boldsymbol{b}\| \|\boldsymbol{a}\|} \le 1\ . \tag{9}$$

Denoted by $C(\boldsymbol{b},\boldsymbol{a}) = |\cos\theta|$ the correlation coefficient between $\boldsymbol{b}$ and $\boldsymbol{a}$. From (8) and (9), we have $0 \leq C(\boldsymbol{b},\boldsymbol{a}) \leq 1$. $\boldsymbol{b}$ and $\boldsymbol{a}$ are orthogonal if $C$ is 0, and $\boldsymbol{b}$ and $\boldsymbol{a}$ are collinear if $C$ is 1.

### 3.1.1 Definition of the proposed ITL-Correlation

Denoted by $\phi(\boldsymbol{b})$ and $\phi(\boldsymbol{a})$ the data mapped in the kernel space, then the angle between $\phi(\boldsymbol{b})$ and $\phi(\boldsymbol{a})$ is actually a kernel version of (8)

$$\theta = \cos^{-1}\left(\frac{\langle\phi(\boldsymbol{b}),\phi(\boldsymbol{a})\rangle}{\|\phi(\boldsymbol{b})\|\|\phi(\boldsymbol{a})\|}\right) = \cos^{-1}\sqrt{2\pi}\sigma k_{\sigma}(\boldsymbol{b},\boldsymbol{a}), \tag{10}$$

where $k_{\sigma}(\cdot\,,\,\cdot)$ is the Gaussian kernel defined in (12). From (10), we know that the Gaussian function calculates the cosine of the angle between two vectors. Since the ITL has a close relationship with the kernel methods, we will introduce a similarity measurement based on the ITL, i.e. correntropy, in the following.

The correntropy between two arbitrary variables $\boldsymbol{b}$ and $\boldsymbol{a}$ is defined by [30-33]

$$V(\boldsymbol{b},\boldsymbol{a}) = \mathrm{E}[k(\boldsymbol{b},\boldsymbol{a})] = \int k(\boldsymbol{b},\boldsymbol{a})dF_{\boldsymbol{ba}}(\boldsymbol{b},\boldsymbol{a}), \tag{11}$$

where $\mathrm{E}(\cdot)$ and $k(\cdot\,,\,\cdot)$ are the mathematical expectation operator and shift-invariant Mercer kernel, respectively. $F_{\boldsymbol{ba}}(\boldsymbol{b},\boldsymbol{a})$ denotes the joint distribution function of $(\boldsymbol{b},\boldsymbol{a})$. In this paper, the Gaussian function is used as the kernel function unless otherwise stated.

$$k(\boldsymbol{b},\boldsymbol{a}) = k_{\sigma}(\boldsymbol{b},\boldsymbol{a}) = \frac{1}{\sqrt{2\pi}\sigma}\exp\left(-\frac{(\boldsymbol{b}-\boldsymbol{a})^2}{2\sigma^2}\right). \tag{12}$$

From the relationship between (10) and (11), we observed that the correntropy defined in (11) can be used as a correlation function to estimate the similarity between two arbitrary vectors $\boldsymbol{b}$ and $\boldsymbol{a}$

$$V(\boldsymbol{b},\boldsymbol{a}) = \frac{1}{\sqrt{2\pi}\sigma}\exp\left(-\frac{\|\boldsymbol{b}-\boldsymbol{a}\|_2^2}{2\sigma^2}\right). \tag{13}$$

However, directly using (13) for computing the correlation between two vectors sometimes fails to estimate the true correlation between $\boldsymbol{b}$ and $\boldsymbol{a}$ since the term $||\boldsymbol{b}-\boldsymbol{a}||_2^2$ in fact estimates the Euclidean distance between two vectors. In sparse representation, given an overcomplete dictionary $\in R^{m\times N}$ , where $m$ and $N$ are the dimension and number of the atoms in $\boldsymbol{A}$, a query sample $\boldsymbol{b}\in R^m$ can be represented by a linear combination of atoms in $\boldsymbol{A}$**.** Assume there is a set of coefficients $\{\beta_1,\beta_2,\dots,\beta_N\}$, the relationship between $\boldsymbol{A}$ and $\boldsymbol{b}$ can be described by $\boldsymbol{b}=\beta_1\boldsymbol{a}_1+\beta_2\boldsymbol{a}_2+\cdots+\beta_N\boldsymbol{a}_N$ , where $\boldsymbol{a}_i$ is the atom in $\boldsymbol{A}$. These descriptions show that $\beta_i$ is not fixed at 1 but can be any scalar. The reason why the term $||\boldsymbol{b}-\boldsymbol{a}||_2^2$ fails to describe a correct relation between $\boldsymbol{b}$ and atoms in $\boldsymbol{A}$ is because it only considers how far $\boldsymbol{a}$ is from $\boldsymbol{b}$ (Euclidean distance) instead of how much each atom contributes to $\boldsymbol{b}$, and this contribution level is well controlled by $\beta_i$. Thus, to correctly characterise this relationship, we propose a new correlation function called ITL-Correlation for correlation measurement between vectors.

***Definition 1*****:** Given vectors $\boldsymbol{b}$ and $\boldsymbol{a}$, the ITL-Correlation is defined by

$$V(\boldsymbol{b},\boldsymbol{a})=\frac{1}{\sqrt{2\pi}\sigma}\exp\left(-\frac{||\boldsymbol{b}-\beta\boldsymbol{a}||_2^2}{2\sigma^2}\right), \tag{14}$$

where $\beta$ is a scalar variable that controls the contribution of $\boldsymbol{a}$ to $\boldsymbol{b}$. The quadratic term $||\boldsymbol{b}-\beta\boldsymbol{a}||_2^2$ can be viewed as a fitting error of the linear regression $\boldsymbol{b}=\beta\boldsymbol{a}+\boldsymbol{v}$, where $\boldsymbol{b}$ and $\boldsymbol{a}$ are linear variables, $\beta$ is the slope of the linear relationship, and $\boldsymbol{v}$ is the offset (fitting error). Since $\beta$ controls the relationship between $\boldsymbol{b}$ and $\boldsymbol{a}$, and further affects the ITL-Correlation in (14), thus to solve $\beta$ and maximize the ITL-Correlation between vectors $\boldsymbol{b}$ and $\boldsymbol{a}$, we minimize the following quadratic error function

$$f(\beta)=\min ||\boldsymbol{b}-\beta\boldsymbol{a}||_2^2, \tag{15}$$

which is a least-squares problem and has a closed-form solution that is unique. By setting $f(\beta)=0$ and solving the least-squares problem, we obtain the unique contribution coefficient $\beta^* = \boldsymbol{a}^T\boldsymbol{b}(\boldsymbol{a}^T\boldsymbol{a})^{-1}$ which indicates that $\boldsymbol{b}$ can be represented by $\boldsymbol{a}$, and the correlation between $\boldsymbol{b}$ and $\boldsymbol{a}$ is maximized.

### 3.1.2 Properties of the proposed ITL-Correlation

Motivated by the analyses on the relationship between two vectors in information theoretic learning [25, 26, 32] and in the $l_p$- space [23, 34], we present some properties of ITL-Correlation in (14) below:

***Property 1:*** If any of the vectors in $\boldsymbol{b}$ and $\boldsymbol{a}$ is a zero vector, then $V(\boldsymbol{b},\boldsymbol{a}) \rightarrow 0$, which means: 1) a zero vector always has zero ITL-Correlation with any other vector; 2) a zero ITL-Correlation also indicates that there is no relationship between $\boldsymbol{b}$ and $\boldsymbol{a}$; 3) a zero ITL-Correlation denotes that the directions of both vectors are different.

***Property 2:*** If $\boldsymbol{b} = \boldsymbol{a}$, then the contribution coefficient $\beta = 1$, and the ITL-Correlation between vectors $\boldsymbol{b}$ and $\boldsymbol{a}$ becomes the correlation of a vector with itself. In this case, we say vector $\boldsymbol{a}$ has the maximum ITL-Correlation with vector $\boldsymbol{b}$, the ITL-Correlation then becomes $V(\boldsymbol{a},\boldsymbol{a})$ or $V(\boldsymbol{b},\boldsymbol{b})$, and $V(\boldsymbol{a},\boldsymbol{a}) = V(\boldsymbol{b},\boldsymbol{b}) = \frac{1}{\sqrt{2\pi}\sigma}$.

***Property 3:*** Given an arbitrary constant $\tau \in R$ and $\tau \neq 1$, the ITL-Correlation has the following properties:

$$\begin{cases} V(\tau\boldsymbol{b},\boldsymbol{a}) = \left(\frac{1}{\sqrt{2\pi}\sigma}\right)^{(1-|\tau|^2)} \left(V(\boldsymbol{b},\boldsymbol{a})\right)^{|\tau|^2}, \\ V(\boldsymbol{b},\tau\boldsymbol{a}) = V(\boldsymbol{b},\boldsymbol{a}), \end{cases} \tag{16}$$

which means that the ITL-Correlation is scale invariant if changing the scale of $\boldsymbol{a}$ by a factor, but is scale variant if changing the scale of $\boldsymbol{b}$ **(**see proof in Appendix A).

***Property 4:*** $V(\boldsymbol{b},\boldsymbol{a})$ is non-negative and bounded. i.e., $0 \leq V(\boldsymbol{b},\boldsymbol{a}) \leq \frac{1}{\sqrt{2\pi}\sigma}$. If there is no relationship between $\boldsymbol{b}$ and $\boldsymbol{a}$, i.e., the optimal contribution coefficient $\beta^*$ is very close to 0, then $V(\boldsymbol{b},\boldsymbol{a})$ reaches its minimum. An increasing $\beta^*$ indicates that $\boldsymbol{b}$ and $\boldsymbol{a}$ have closer relationship and $V(\boldsymbol{b},\boldsymbol{a})$ reaches its maximum when $\boldsymbol{b} - \beta^*\boldsymbol{a} = \boldsymbol{0}$ $(\boldsymbol{b} = \beta^*\boldsymbol{a})$, which means that $\boldsymbol{b}$ can be represented by $\boldsymbol{a}$ under a proper $\beta^*$ and this $\beta^*$ is unique. According to the linear algebra theory [35], we give the proof in Appendix B.

### 3.1.3 Normalized ITL-Correlation

Scale invariance is very important in correlation measurement. If we convert the original variables $\boldsymbol{b}$ and $\boldsymbol{a}$ to the new variables $\tilde{\boldsymbol{b}}$ ($\tilde{\boldsymbol{b}} = \tau \boldsymbol{b}$) and $\tilde{\boldsymbol{a}}$ ($\tilde{\boldsymbol{a}} = \tau \boldsymbol{a}$) by an arbitrary scale factor $\tau$, then the distribution of the original variable may not be preserved in the new variable because the scale change on a variable is likely to lead to change of distribution. Normalization is a common tool for rescaling a vector, avoiding the variance of a vector for a correlation metric or a similarity measure [23]. Since the invariance of the ITL-Correlation with respect to the variable $\boldsymbol{b}$ directly controls the invariability of the correlation, here we introduce a normalization factor $||\boldsymbol{b}||_2^2$ to avoid the variance caused by rescaling $\boldsymbol{b}$. The resulted normalized ITL-Correlation will be invariant with respect to vectors $\boldsymbol{b}$ and $\boldsymbol{a}$ at the same time.

***Definition 2:*** Denoted by $V_N(\boldsymbol{b}, \boldsymbol{a})$ the normalized ITL-Correlation, the definition is given by

$$V_N(\boldsymbol{b}, \boldsymbol{a}) = \frac{1}{\sqrt{2\pi}\sigma} \exp\left(-\frac{||\boldsymbol{b} - \beta \boldsymbol{a}||_2^2}{2\sigma^2 ||\boldsymbol{b}||_2^2}\right). \tag{17}$$

The properties of the normalized ITL-Correlation in (17) is similar to that in (14), which are given as follows:

***Property 1:*** If $\boldsymbol{b}$ or $\boldsymbol{a}$ is a zero vector, then $V_N(\boldsymbol{b}, \boldsymbol{a}) \to 0$, which means that a zero vector always has a zero normalized ITL-Correlation with any vector.

***Property 2:*** If $\boldsymbol{b} = \boldsymbol{a}$, vector $\boldsymbol{a}$ has the maximum normalized ITL-Correlation with vector $\boldsymbol{b}$, i.e., $V_N(\boldsymbol{b}, \boldsymbol{a}) = \frac{1}{\sqrt{2\pi}\sigma}$.

***Property 3:*** The normalized ITL-Correlation is scale invariant with respect to both vectors $\boldsymbol{b}$ and $\boldsymbol{a}$, i.e., $V_N(\boldsymbol{b}, \tau \boldsymbol{a}) = V_N(\boldsymbol{b}, \boldsymbol{a})$, and $V_N(\tau \boldsymbol{b}, \boldsymbol{a}) = V_N(\boldsymbol{b}, \boldsymbol{a})$.

***Property 4:*** $V_N(\boldsymbol{b},\boldsymbol{a})$ is non-negative and bounded, i.e., $0 \leq V_N(\boldsymbol{b},\boldsymbol{a}) \leq \frac{1}{\sqrt{2\pi}\sigma}$.

### 3.1.4 ITL-Orthogonality

In orthogonal matching pursuit algorithms, vectors $\boldsymbol{b}$ and $\boldsymbol{a}$ are orthogonal if they are uncorrelated. Here we evaluate the ITL-Orthogonality of our proposed method.

***Definition 3:*** If $V(\boldsymbol{b},\boldsymbol{a}) = 0$, then any two different arbitrary vectors $\boldsymbol{b}$ and $\boldsymbol{a}$ are orthogonal. Based on above definitions and properties of ITL-Correlation, we know that a vector is ITL-Orthogonal to another vector if these two vectors has a zero ITL-Correlation with each other. Some important properties of ITL-Orthogonality are given as follows:

***Property 1:*** If $\beta = 0$, then $V(\boldsymbol{b},\boldsymbol{a}) \to 0$, and $\boldsymbol{b}$ and $\boldsymbol{a}$ are ITL-Orthogonal.

***Property 2:*** If $\boldsymbol{b}$ and $\boldsymbol{a}$ are nonzero vectors and $V(\boldsymbol{b},\boldsymbol{a}) = 0$, then we have $\beta = 0$.

### 3.1.5 Proposed algorithm for computing ITL-Correlation

Given a dictionary $\boldsymbol{A}$ containing $N$ training samples and a testing sample $\boldsymbol{b}$, every dictionary atom is a vectorized sample of dimension $m$, forming the dictionary $\boldsymbol{A} = [\boldsymbol{a}_1, \boldsymbol{a}_2, \cdots, \boldsymbol{a}_N] \in R^{m \times N}$. The testing sample $\boldsymbol{b}$ is also represented as a vectorized sample of dimension $m$. In order to select an atom which makes the largest contribution in representing the testing sample $\boldsymbol{b}$, we need to calculate the ITL-Correlation between each dictionary atom $\boldsymbol{a}_i = [a_1, a_2, \cdots, a_m]^T \in R^m$ and $\boldsymbol{b}$. To calculate the ITL-Correlation defined by (14), we need to solve (15) for the optimal weight coefficient $\beta^*$. Since (15) is a closed-form function, we can estimate $\beta^*$ by solving the least squares problem

$$\beta^* = \arg\min_{\beta} \|\boldsymbol{b} - \beta \boldsymbol{a}_i\|_2^2 = \frac{\boldsymbol{a}_i^T \boldsymbol{b}}{\boldsymbol{a}_i^T \boldsymbol{a}_i}. \tag{18}$$

Then the kernel width parameter can be updated by

$$\sigma^2 = \frac{1}{2m} \|\boldsymbol{b} - \beta^* \boldsymbol{a}_i\|_2^2. \tag{19}$$

Finally, the ITL-Correlation between each dictionary atom $\boldsymbol{a}_i$ and the testing sample $\boldsymbol{b}$ can be calculated by

$$V(\boldsymbol{b}, \boldsymbol{a}_i) = \frac{1}{\sqrt{2\pi}\sigma} \exp\left(-\frac{\|\boldsymbol{b} - \beta^* \boldsymbol{a}_i\|_2^2}{2\sigma^2}\right). \tag{20}$$

After obtaining all the correlations corresponding to all the dictionary atoms, the ITL-Correlation coefficients are $V = \{V(\boldsymbol{b}, \boldsymbol{a}_1), V(\boldsymbol{b}, \boldsymbol{a}_2), \cdots, V(\boldsymbol{b}, \boldsymbol{a}_N)\} \in R^{1\times N}$. To find which dictionary atom most correlates with the testing sample, we select the index of the largest value in $V$:

$$\lambda = \underset{i=1,2,\cdots,N}{\arg\max} \{V(\boldsymbol{b}, \boldsymbol{a}_i)\}. \tag{21}$$

The proposed ITL-Correlation algorithm is summarized in Algorithm 1.

---

**Algorithm 1** Computation of ITL-Correlation

---

**Input:** Sample matrix $\boldsymbol{A} = [\boldsymbol{a}_1, \boldsymbol{a}_2, \cdots, \boldsymbol{a}_N] \in R^{m\times N}$ with each vector $\boldsymbol{a}_i = [a_1, a_2, \cdots, a_m]^T \in R^m$, and $\boldsymbol{b} = [b_1, b_2, \cdots, b_m]^T \in R^m$.

**Output:** The index λ of the largest correlation coefficient.

1: **for** $i = 1,2,\cdots,N$ **do**

2: Calculate the optimal weight coefficient $\beta^*$ using (18).

3: Calculate the kernel width parameter σ by (19).

4: Calculate the ITL-Correlation coefficient $V(\boldsymbol{b}, \boldsymbol{a}_i)$ by (20).

5: **end for**

6: The index of the dictionary atom that most correlates with the testing sample can be calculated by (21)

---

## 3.2. Non-Second Order Kernel Minimization

### 3.2.1 OMP Algorithm Based on ITL-Correlation and Non-Second Order Kernel Statistic Measurement (INOK-OMP)

In this section, we apply the ITL-Correlation and correntropy loss concepts to the orthogonal matching pursuit algorithm to develop a new robust matching pursuit algorithm. Given a training dataset $\boldsymbol{A} = \left[\boldsymbol{A}_{C_1}, \boldsymbol{A}_{C_2}, \cdots, \boldsymbol{A}_{C_i}, \cdots, \boldsymbol{A}_{C_K}\right] = [\boldsymbol{a}_1, \boldsymbol{a}_2, \cdots, \boldsymbol{a}_N] \in R^{m\times N}$ and a testing sample $\boldsymbol{b}$, then the testing image can be represented by a linear combination of training samples in $\boldsymbol{A}$. Here $\boldsymbol{A}_{C_i}$ means the training samples from the $C_i$th class, and $K$ is the number of the classes. Before implementing the INOK-OMP algorithm, the supporting set and residual need to be initialized as $S = \Phi$ and $\boldsymbol{r}_0 = \boldsymbol{b}$, respectively. For a $k$th iteration, we find an atom that has the highest ITL-Correlation with the current residual in $\boldsymbol{A}$, then add the index of this atom to the support set. Then the new sparse coefficient $\boldsymbol{x}_k$ can be updated by minimizing the following NOK-loss function

$$
\begin{aligned}
J_{\text{NOK-loss}}(\boldsymbol{x}) &= \frac{1}{m}\sum_{j=1}^{m}\left(1 - k_\sigma\left(\boldsymbol{b}_j - \sum_{i=1}^{N}\boldsymbol{a}_{ij}\boldsymbol{x}_i\right)\right)^{\frac{p}{2}} \\
&= \frac{1}{m}\sum_{j=1}^{m}\left(1 - k_\sigma(\boldsymbol{e}_j)\right)^{\frac{p}{2}} \\
&= \frac{1}{m}\sum_{j=1}^{m}\rho\left(\left\|\boldsymbol{e}_j\right\|_2\right), \qquad (22)
\end{aligned}
$$

where $\boldsymbol{e}_j = \boldsymbol{b}_j - \sum_{i=1}^{N}\boldsymbol{a}_{ij}\boldsymbol{x}_i$ , and $\rho\left(\left\|\boldsymbol{e}_j\right\|_2\right) = (1 - k_\sigma(\boldsymbol{b}_j - \sum_{i=1}^{N}\boldsymbol{a}_{ij}\boldsymbol{x}_i))^{\frac{p}{2}}$ . It is easy to check that the function $\rho\left(\left\|\boldsymbol{e}_j\right\|_2\right)$ has properties: i) $\rho\left(\left\|\boldsymbol{e}_j\right\|_2\right) \geq 0$; ii) $\rho(0) = 0$; iii) $\rho\left(\left\|\boldsymbol{e}_j\right\|_2\right) = \rho\left(-\left\|\boldsymbol{e}_j\right\|_2\right)$; and iv) $\rho\left(\left\|\boldsymbol{e}_{j1}\right\|_2\right) > \rho\left(\left\|\boldsymbol{e}_{j2}\right\|_2\right)$, for $\left\|\boldsymbol{e}_{j1}\right\|_2 > \left\|\boldsymbol{e}_{j2}\right\|_2$. Thus, $\rho\left(\left\|\boldsymbol{e}_j\right\|_2\right)$ belongs to the $M$-estimation type of robust cost function, and minimizing the

loss in (22) is an $M$-estimation problem [30, 36]. Based on the theory of $M$-estimation, the minimization of (22) can be transformed into a re-weighted least-squares problem which has already been successfully applied in the field of computer vision [37-40]. Based on the theory of $M$-estimation, we solve the problem in (22) by minimizing the re-weighted least-squares function as follows.

$$\min \sum_{j=1}^{m} \rho\left(\|\boldsymbol{e}_j\|_2\right) = \min \sum_{j=1}^{m} \gamma\left(\|\boldsymbol{e}_j\|_2\right)\|\boldsymbol{e}_j\|_2^2, \tag{23}$$

where function $\gamma(\|\boldsymbol{e}_j\|_2)$ is defined by

$$\gamma_j(\|\boldsymbol{e}_j\|_2) = \frac{\rho'\left(\|\boldsymbol{e}_j\|_2\right)}{\|\boldsymbol{e}_j\|_2}, \tag{24}$$

in which $\rho'\left(\|\boldsymbol{e}_j\|_2\right)$ is the derivative of $\rho\left(\|\boldsymbol{e}_j\|_2\right)$. Therefore, the weight in the $k$th iteration is calculated by

$$\gamma_k^j = \frac{p}{2\sigma_k^2}\left[1 - \exp\left(-\frac{\|\boldsymbol{e}_j\|_2^2}{2\sigma_k^2}\right)\right]^{\frac{p}{2}-1} \exp\left(-\frac{\|\boldsymbol{e}_j\|_2^2}{2\sigma_k^2}\right), \qquad j = 1, \cdots, m, \tag{25}$$

where $\sigma_k$ is

$$\sigma_k = \sqrt{\frac{1}{2m}\|\boldsymbol{b} - \boldsymbol{A}_{S_k}\boldsymbol{x}_{k-1}\|_2^2} \tag{26}$$

and $\boldsymbol{A}_{S_k}$ are the selected atoms corresponding to the supporting set $S_k$. From (25), we know that a larger representation error produces smaller weights, which means that the learned features are mainly dominated by the contribution of inliers not the outliers after a suitable number of iterations, thus the NOK-loss based function is robust against outliers. Based on the discriminative weights $\boldsymbol{\gamma}_k = \{\gamma_k^1, \cdots, \gamma_k^m\}$ learned by (25), the sparse coefficient can be updated by

$$\boldsymbol{x}_k = \underset{x \in R^N, \mathrm{supp}(\boldsymbol{x}) \subset S_k}{\operatorname{argmin}} \left\|\sqrt{\mathrm{diag}(\boldsymbol{\gamma}_k)}\left(\boldsymbol{b} - \boldsymbol{A}_{S_k}\boldsymbol{x}^{k-1}\right)\right\|_2^2, \tag{27}$$

**Algorithm 2.** INOK-OMP algorithm

**Input:** Dictionary $\boldsymbol{A} = [\boldsymbol{a}_1, \boldsymbol{a}_2, \cdots, \boldsymbol{a}_N] \in R^{m\times N}$ with $\boldsymbol{a}_i = [a_1, a_2, \cdots, a_m]^T \in R^m$, test sample $\boldsymbol{b} = [b_1, b_2, \cdots, b_m]^T \in R^m$, error threshold $\epsilon$.

**Output:** Sparse vector $\boldsymbol{x}_k$.

**Initialization:** $k = 0$, the initial sparse vector $\boldsymbol{x}_0 = \mathbf{1}$, the initial residual $\boldsymbol{r}_0 = \boldsymbol{b}$, the initial solution support $S_0 = \text{Support}\{\boldsymbol{x}_0\} = \Phi$.

1: **while do**

2: **Sweep:** find an atom that has the highest ITL-Correlation with the current residual in $\boldsymbol{A}$ using Algorithm 1.

$$\lambda_k = \underset{i=1,2,\cdots,N}{\arg\max}\,\{V(\boldsymbol{r}_k, \boldsymbol{a}_i)\}.$$

3: **Update support:** update the support $S_k = S_{k-1} \cup \{\lambda_k\}$.

4: **Update solution:** Calculate $\boldsymbol{x}_k$ by minimizing $\left\|\boldsymbol{b} - \boldsymbol{A}_{S_k}\boldsymbol{x}_{k-1}\right\|_2^2$ subject to $\text{Support}\{\boldsymbol{x}_{k-1}\} = S_k$. Update $\sigma_k$ using (26), update $\gamma_k^j$ using (25), update $\boldsymbol{x}_k$ using (27). The optimal solution is denoted as $(\boldsymbol{\gamma}_k, \boldsymbol{x}_k)$.

5: **Update residual:** $\boldsymbol{r}_{k+1} = \sqrt{\text{diag}(\boldsymbol{\gamma}_k)}\left(\boldsymbol{b} - \boldsymbol{A}_{S_k}\boldsymbol{x}_k\right)$.

6: **Until** $\|\boldsymbol{r}_{k+1}\|_2 < \epsilon$

7: **end while**

which is a closed-form function, and can be solved by

$$\boldsymbol{x}_k = \frac{\widehat{\boldsymbol{A}}^T\widehat{\boldsymbol{b}}}{\widehat{\boldsymbol{A}}^T\widehat{\boldsymbol{A}}}\,, \tag{28}$$

where $\widehat{\boldsymbol{A}} = \sqrt{\text{diag}(\boldsymbol{\gamma}_k)}\boldsymbol{A}_{S_k}$, and $\widehat{\boldsymbol{b}} = \sqrt{\text{diag}(\boldsymbol{\gamma}_k)}\,\boldsymbol{b}$. The residual can be updated based on the new weight $\boldsymbol{\gamma}_k$ and sparse vector $\boldsymbol{x}_k$ by

$$\boldsymbol{r}_{k+1} = \sqrt{\text{diag}(\boldsymbol{\gamma}_k)}(\boldsymbol{b} - \boldsymbol{A}_{S_k}\boldsymbol{x}_k). \tag{29}$$

Finally, the proposed minimization of NOK-loss in (22) can be solved using the alternate procedure from (25)-(29). The INOK-OMP algorithm is summarized in Algorithm 2.

### 3.2.2 Learning Robust Classifier with NOK Statistic Measurement

In sparse representation-based classification, the validity of classifying samples largely depends on the reconstruction error from a specific class. Thus, how to design an effective criterion to calculate the reconstruction error becomes a key issue. Unlike other research work that designs the classifier based on the Euclidean distance [1, 41] or correntropy [22, 24] of the reconstruction error, we design the classifier based on the proposed NOK statistic measurement. Let $\delta_{C_i}(\boldsymbol{x})$ be the sparse coefficients corresponding to the $C_i, i = 1, \cdots, K$. For each class, we obtain the approximated representation as $\widehat{\boldsymbol{b}}_{C_i} = \boldsymbol{A}_{C_i}\delta_{C_i}(\boldsymbol{x})$. Then based on the NOK-loss, the query sample $\boldsymbol{b}$ can be classified into a class $i$ which leads the minimal difference between $\boldsymbol{b}$ and $\widehat{\boldsymbol{b}}_{C_i}$. The detailed robust NOK-classifier is described in Algorithm 3.

---

**Algorithm 3.** Robust NOK-classifier

---

**Input:** Sample matrix $\boldsymbol{A} = [\boldsymbol{a}_1, \boldsymbol{a}_2, \cdots, \boldsymbol{a}_N] \in R^{m\times N}$ with each vector $\boldsymbol{a}_i = [a_1, a_2, \cdots, a_m]^T \in R^m$, a test sample $\boldsymbol{b} = [b_1, b_2, \cdots, b_m]^T \in R^m$.

**Output:** Identity$(\boldsymbol{b})$

1: Solve the sparse coefficients $\boldsymbol{x}$ of test sample $\boldsymbol{b}$ using Algorithm 2.

2: Calculate the kernel width parameter σ by $\sigma^2 = \frac{1}{2m}||\boldsymbol{b} - \boldsymbol{A}\boldsymbol{x}||_2^2$

3: Calculate the residual of each class using

$$\boldsymbol{r}_{C_i}(\boldsymbol{b}) = \left(1 - k_\sigma\left(\boldsymbol{b} - \boldsymbol{A}_{C_i}\delta_{C_i}(\boldsymbol{x})\right)\right)^{\frac{p}{2}}, \text{for } i = 1,2,\cdots,K$$

4: Identity$(\boldsymbol{b})$ = argmin$_{C_i}$ $\boldsymbol{r}_{C_i}(\boldsymbol{b})$.

---

## 3.3. Computational Complexity Analysis

As with most OMP algorithms, the proposed method consists of three parts, including identification, estimation, and residual update. The computational complexity of each part of the proposed algorithm is analyzed as follows.

1) **Identification**: for the identification part, the complexity comes from computing the ITL-Correlation, which consists of four steps of computation for contribution coefficients $\beta^+$, kernel width $\sigma^2$, ITL-Correlation coefficients $V$, and the maximum $\lambda$. $\beta^+$ can be obtained by $\frac{\boldsymbol{A}^T\boldsymbol{b}}{\boldsymbol{A}^T\boldsymbol{A}}$ with a complexity of $MN + N^3$ flops, where $M = 2m - 1$, $m$ is the dimension of $\boldsymbol{a}_i$ and $\boldsymbol{b}$, $N$ is the size of $\boldsymbol{A}$, and $N^3$ is the complexity of the matrix inversion operation $(\boldsymbol{A}^T\boldsymbol{A})^{-1}$. Here $\beta^+$ is a vector that consists of all the contribution coefficients corresponding to different atoms in $\boldsymbol{A}$. Computation of $\sigma^2$ and $V$ needs $MN$ flops each. The identification part also needs $N$ flops for finding the maximum in $V$. Thus, the total computational complexity of the identification part is $MN + N^3 + MN + MN + N$ flops.
2) **Estimation**: this part consists of three steps for computing kernel width $\sigma$, weight $\gamma$, and sparse coefficient $\boldsymbol{x}_k$, respectively. We need $MK$ flops to obtain $\sigma$, $MK + MK$ flops to obtain $\gamma$, and $K^3 + MK$ flops to obtain $\boldsymbol{x}_k$, where $K$ is the size of the support set. Since this estimation part executes in an alternating manner, the total complexity of estimation is $(MK + (MK + MK) + (K^3 + MK))l$, where $l$ is the number of the iterations.
3) **Residual update**: to obtain the residual $r$, we need $(2K - 1)m$ flops.

Table 1 compares the computational complexity of the proposed algorithm and those of benchmarks, including OMP [11], GOMP [12], CoSaMP [13], CMP [24], LOMP [17], NR-A*OMP [15], Acc-MMP [16], and KNS-OMP [27].

For consistency, we use the same method as [12] to compute the complexity of different algorithms in this table. For example, $\boldsymbol{a}_i^T\boldsymbol{b}$ needs $m$ flops for multiplication and $(m-1)$ flops for addition. Thus, the total complexity of $\boldsymbol{a}_i^T\boldsymbol{b}$ is $M$ flops, and $MK$ flops for $\boldsymbol{A}^T\boldsymbol{b}$, From this table, we can see that the computational complexity of the proposed algorithm is dominated by the estimation part which uses an alternating manner for optimization. For the computational complexity of LOMP [17], $Z$ is the number of hidden unites.

Table 1 Computational complexity of different algorithms.

| Methods | Computational Complexity | | |
|---|---|---|---|
| | Identification | Estimation | Residual Update |
| OMP | $MN+N$ | $MK+K^3$ | $(2K-1)m$ |
| GOMP | $MN+\frac{NI-N(N+1)}{2}$ | $4I^2Km+(-2I^2+5I)m+2I^3K^2+(-4I^3+5I^2)K+3I^3-I^2-I$ | $(2K-1)m$ |
| CoSaMP | $MN+N$ | $K+mK$ | $(2K-1)m$ |
| CMP | $MN+N$ | $(MK+MK\ +K^3+MK)l$ | $(2K-1)m$ |
| LOMP | $ZN+Z^3+NZ^3d+NZd$ | | |
| NR-A*OMP | The complexity of this algorithm depends on the number of iterations, paths, and the pruning strategies introduced. Its complexity with different parameters varies between that of an exhaustive search and that of a simple OMP algorithm. | | |
| Acc-OMP | The complexity of this algorithm depends on a tree search strategy, and is higher than that of OMP. This special tree search scheme provides a balance between the computational complexity and the performance. | | |
| KNS-OMP | $MN+N$ | $(MK+(MK+MK)\ +(K^3+MK))l$ | $(2K-1)m$ |
| Proposed | $MN+N^3+MN+MN+N$ | $(MK+(MK+MK)\ +(K^3+MK))l$ | $(2K-1)m$ |

# 4. Experimental Results

To verify the effectiveness of the proposed INOK-OMP algorithm, we carry out experiments on both synthetic data for different data recovery and real-world image for reconstruction and classification tasks, comparing its performance across various evaluation measures, and comparing it to benchmarks including OMP [11], GOMP [12], CoSaMP [13], CMP [24], LOMP [17], NR-A*OMP [15], Acc-MMP [16], and the KNS-OMP [27].

## 4.1. Data Recovery on Synthetic Data

At first, we carry out experiments on synthetic noisy data to verify the recovery ability of the proposed algorithm. To simulate the dictionary, we generate a matrix $\boldsymbol{A} \in R^{200\times400}$ in which all the elements are independent Gaussian random variables with zero-mean and unit-variance. To simulate a sparse coefficient, we first generate a zero vector $\boldsymbol{x} \in R^{400\times1}$ and then randomly set ten entries in $\boldsymbol{x}$ to random values in $[-2, -1] \cup [1, 2]$. Then the noisy data $\boldsymbol{b}$ can be represented by a linear combination of atoms in $\boldsymbol{A}$ with coefficients corresponding to values in $\boldsymbol{x}$, i.e., $\boldsymbol{b} = \boldsymbol{Ax} + \boldsymbol{n}$ , where $\boldsymbol{n}$ denotes the noise vector.

In this experiment, we mainly consider two categories of noise, i.e., non-Gaussian noise and Gaussian noise with each category containing several different types of noise. Non-Gaussian noise is identified as: (1) the $\chi^2$ distribution with 1 degree of freedom, (2) the exponential distribution with the mean being 1, and (3) the $t$-distribution with 3 degrees of freedom. Gaussian noise is categorised as: (1) Gaussian noise $N(0, 0.5)$ with zero mean and standard deviation being 0.5, (2) white Gaussian noise (WGN) with its peak signal to noise ratio (SNR) being 2. In addition, experiment with missing data in $\boldsymbol{b}$ is also considered to verify the robustness of the proposed algorithm.

The task of data recovery in sparse representation is to approximate the clean data from noisy observations. In this experiment, we recovery sparse vector $\boldsymbol{x}$ with the given noisy vector $\boldsymbol{b}$. The recovery error is estimated by the Euclidean distance between the estimated sparse vector and the ground truth. Table 2 illustrates the recovery errors of the proposed algorithm and all the benchmarks on the synthetic data with different types of noise. To reduce the deviation of the results, the results of all the algorithms are reported over 20 random trials. We set the percentage of missing data to 10% for the case with missing data in this experiment. The results in Table 2 show that the proposed method always obtains the lowest average recovery error under different types of noise. Moreover, most of the standard deviation values of the recovery errors from the proposed algorithm also remain as the lowest ones in the presence of different types of noise.

Table 2: Average recovery errors of all competing algorithms under various types of noises. Ave. and Std. denote the average result and standard deviation, respectively. Best results are marked in bold. Exp., $t$-dis., MD and WGN are abbreviations of exponential, $t$-distribution, missing data, white Gaussian noise.

| Methods | | Different types of noises | | | | | |
|---|---|---|---|---|---|---|---|
| | | $\chi^2(1)$ | Exp. | $t$-dis. | MD | Gaussian | WGN |
| OMP | Ave. | 14.6988 | 12.8864 | 14.6016 | 0.2379 | 4.5224 | 3.0855 |
| | Std. | 1.6159 | 1.5647 | 1.3991 | 0.0944 | **0.2359** | 0.3285 |
| GOMP | Ave. | 14.4463 | 12.2016 | 14.0891 | 0.2667 | 4.5666 | 3.1036 |
| | Std. | 1.6277 | 1.5447 | 1.2757 | 0.1333 | 0.2529 | 0.3178 |
| COSaMP | Ave. | 16.4872 | 13.5687 | 15.8897 | 0.2831 | 5.0876 | 3.4002 |
| | Std. | 1.4668 | 1.2983 | 2.2077 | 0.1194 | 0.2760 | 0.3386 |
| CMP | Ave. | 6.2882 | 9.2243 | 10.7400 | 0.0052 | 4.8926 | 3.2916 |
| | Std. | 0.7100 | 1.0332 | 1.0988 | 0.0016 | 0.3977 | 0.3256 |
| LOMP | Ave. | 16.4721 | 13.6791 | 15.5109 | 0.3430 | 4.6659 | 3.1250 |
| | Std. | 2.8379 | 1.2327 | 2.2094 | 0.1078 | 0.9334 | 0.6916 |
| NR-A*OMP | Ave. | 16.7187 | 13.8771 | 15.5638 | 0.3215 | 4.9343 | 3.2436 |
| | Std. | 2.6466 | 1.4505 | 2.4312 | 0.1025 | 0.8778 | 0.7567 |
| Acc-MMP | Ave. | 15.9708 | 13.5334 | 15.4279 | 0.3107 | 4.8852 | 3.1751 |
| | Std. | 1.6419 | 1.5366 | 2.4638 | 0.1322 | 0.9180 | 0.7603 |
| KNS-OMP | Ave. | 2.3908 | 2.7334 | 3.1291 | 2.8932e-05 | 2.6675 | 2.1159 |
| | Std. | 0.4516 | **0.6175** | 0.9448 | 1.4039e-05 | 0.5648 | 0.3570 |
| Proposed | Ave. | **2.2757** | **2.5522** | **2.6863** | **9.5115e-09** | **2.2797** | **1.9419** |
| | Std. | **0.4343** | 0.8478 | **0.6372** | **4.4421e-09** | 0.4880 | **0.0987** |

To verify the outlier resistance ability of the proposed algorithm, Fig. 1 plots the recovered signals of different algorithms from the data with corruptions and outliers. All the experiments are implemented on the data corrupted by WGN with outliers embedded in the data. In this paper, outliers with magnitude of $\pm 30\sigma_g$ are added to randomly selected elements in $\boldsymbol{b}$. In all the experiments, the number of the non-zero elements in sparse coefficients is set to 10 ($L = 10$). The results in Fig. 1 show the signal recovery performance of all the algorithms when the SNR of WGN is 10dB, and the number of outliers is 6. The results show that the recovery performances of the ITL-based algorithms are better than those of non-ITL based ones. Although the CMP algorithm [24] and the KNS-OMP algorithm [27] can correctly estimate the position of non-zeros elements of the sparse vector, there still exists recovery errors. The proposed algorithm with $p = 1.7$ can not only accurately estimate the position of the non-zeros in the sparse vector, but also minimize the recovery error at the same time.

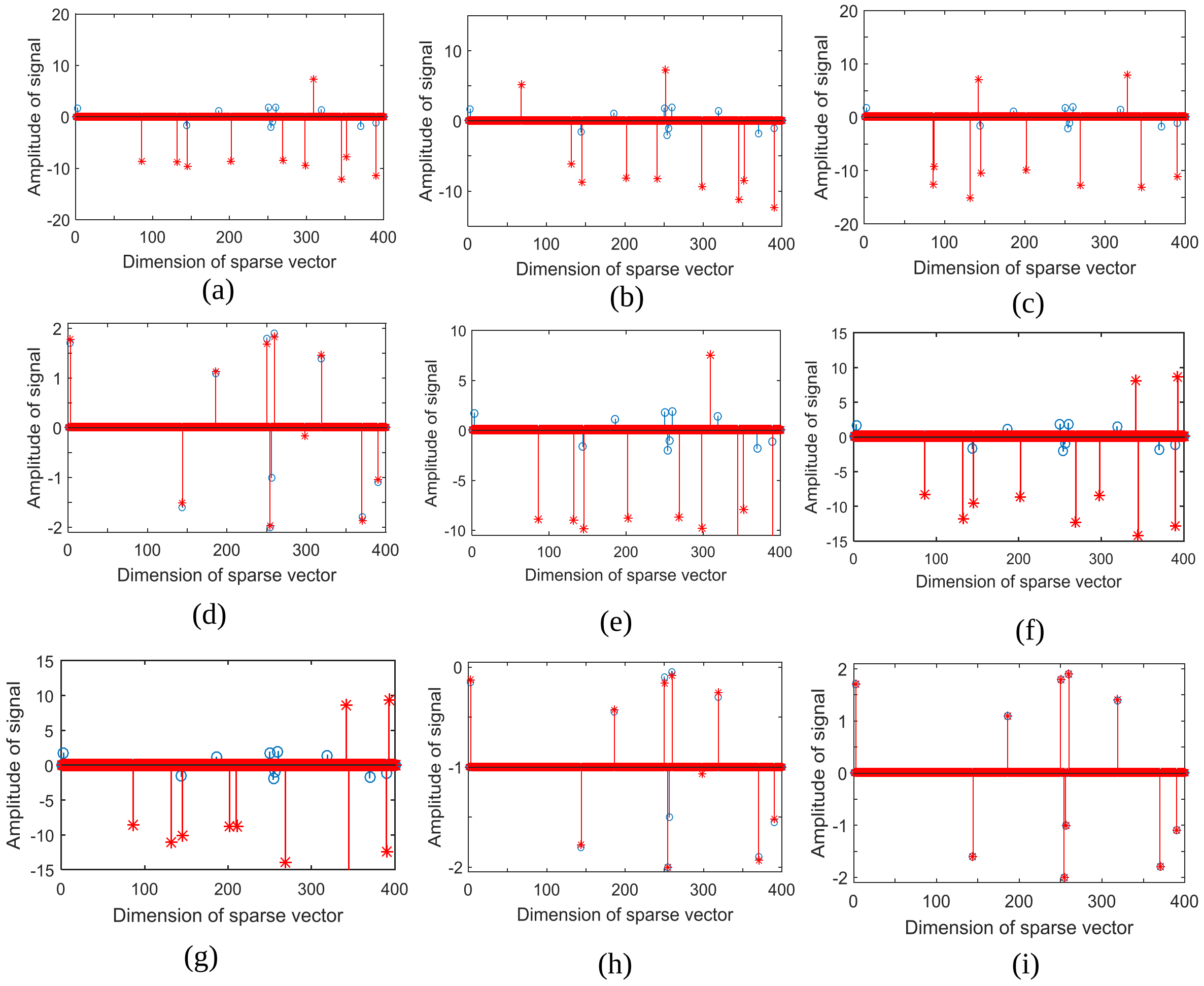

Fig. 1. Recovered sparse vector in the presence of white Gaussian noise with SNR=10dB. The blue/red lines and marks denote the true/recovered signals. (a) OMP, (b) GOMP, (c) CoSaMP, (d) CMP, (e) LOMP, (f) NR-A*OMP, (g) Acc-MMP, (h) KNS-OMP ($p$=1.7), (i) Proposed ($p$ =1.7).

Since we developed a NOK-loss in section 3 to take the advantage of the non-second order measurement in minimizing the representation errors, we here implement experiments to verify the effectiveness of the proposed algorithm with this advantage under different $p$ values. We plot the recovery errors with different types of noise in Fig. 2 where Figs. (a)-(c) are the results from the data with non-Gaussian noise, which show that the recovery errors of the proposed algorithm with $p < 2$ are lower than that with $p = 2$. Although the results from the Gaussian noise cases (e)-(f) are not as good as the non-Gaussian noise cases, it still verifies that the results with $p < 2$ are better than that with $p = 2$. The result of the experiment with missing data is shown in Fig. 2(d) with a magnified part confirming the conclusions of the non-Gaussian and Gaussian noise cases.

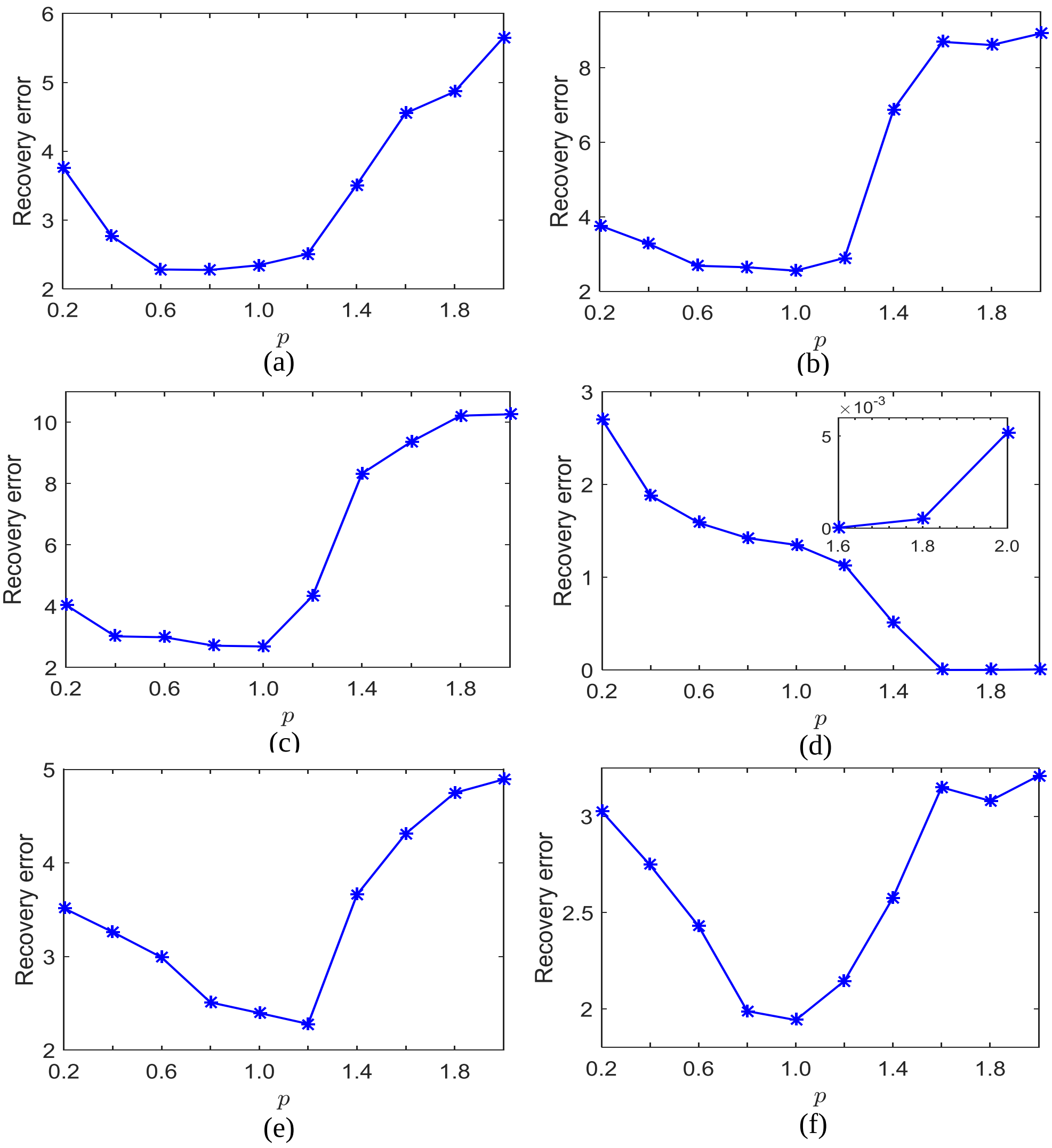

Fig. 2. Average recovery error of the proposed algorithm under different $p$ values with different types of noise. (a) $\chi^2(1)$ distribution, (b) exponential distribution, (c) $t$-distribution, (d) missing data, (e) Gaussian noise, (f) white Gaussian noise.

## 4.2. Image Reconstruction

We now verify the image reconstruction ability of the proposed algorithm on real images with occlusions. All experiments are performed on the extended Yale B face database [42, 43] in which there are 38 subjects with 64 frontal face images being taken under varying illuminations. 32 images per subject are randomly selected to construct the dictionary $\boldsymbol{A}$. Two different types of occlusions are considered in this experiment: one is the randomly generated block with elements varying from 0 to 1, and the other is the sub-block using the real image. To create the corrupted testing images, randomly generated block with size $40 \times 60$ and real image block with size $50 \times 50$ are considered as occlusions for the selected 32 images of the subject. Fig. 3 shows an example original testing image and its corresponding corrupted images.

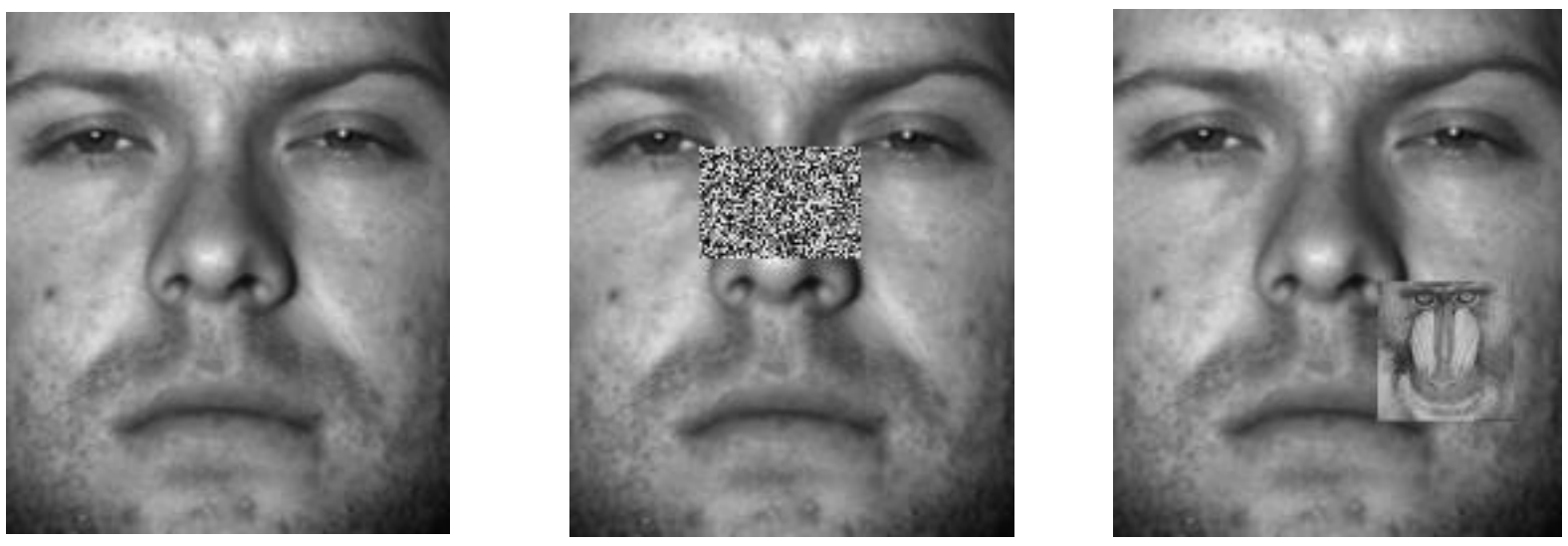

(a) (b) (c)

Fig. 3. Sample images from the extended Yale B dataset. (a) original, (b) randomly generated block occlusion, (c) real image block occlusion.

We quantitatively examine the reconstruction error of the proposed algorithm in comparison with those of benchmarks. All the tests of different algorithms are repeated ten times. The average reconstruction errors of the nine competing algorithms on images from all 38 classes (i.e., subjects) are reported in Fig. 4. The results show that the reconstruction error of the proposed algorithm remains the lowest consistently across all classes under both random and real occlusion cases.

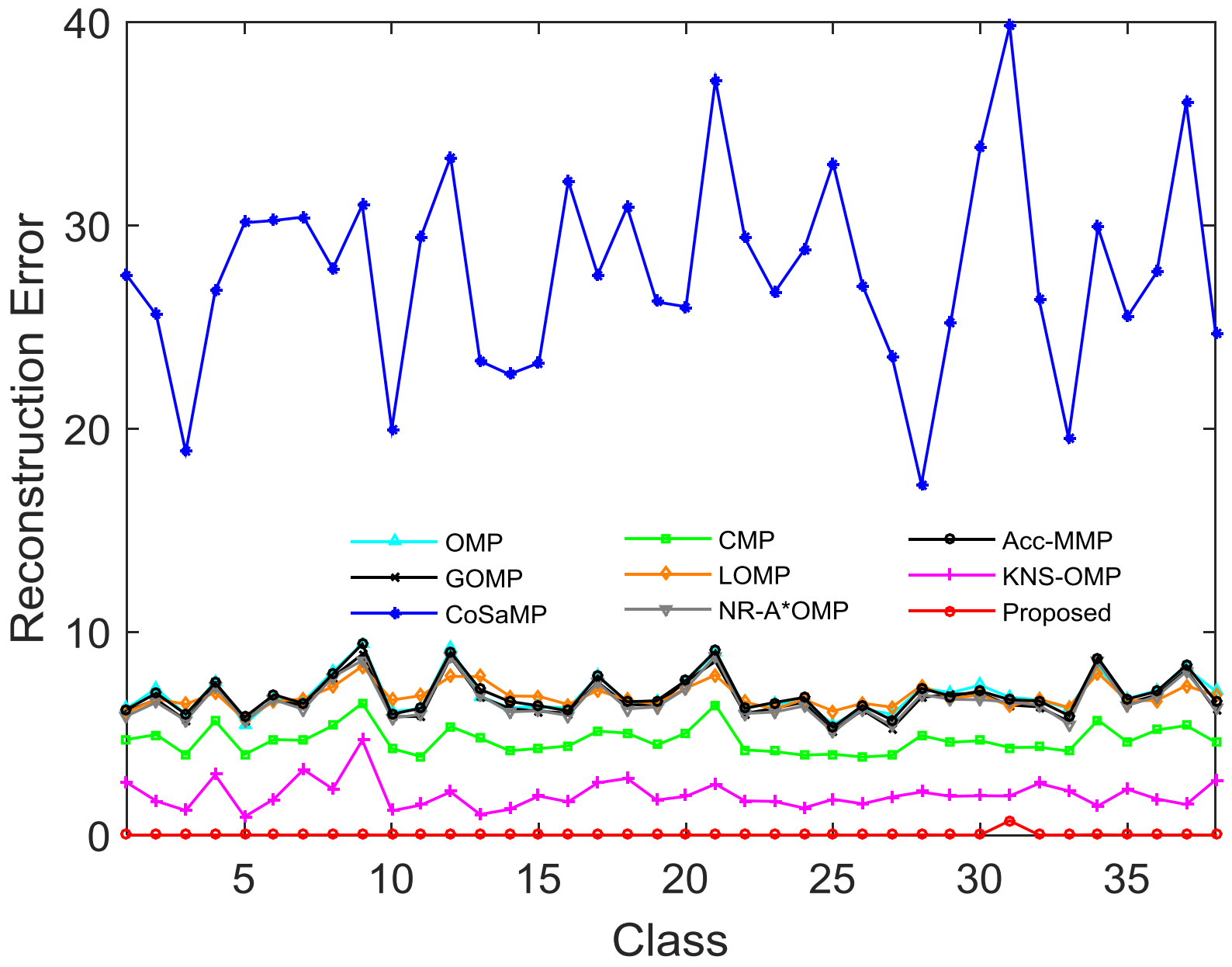

(a)

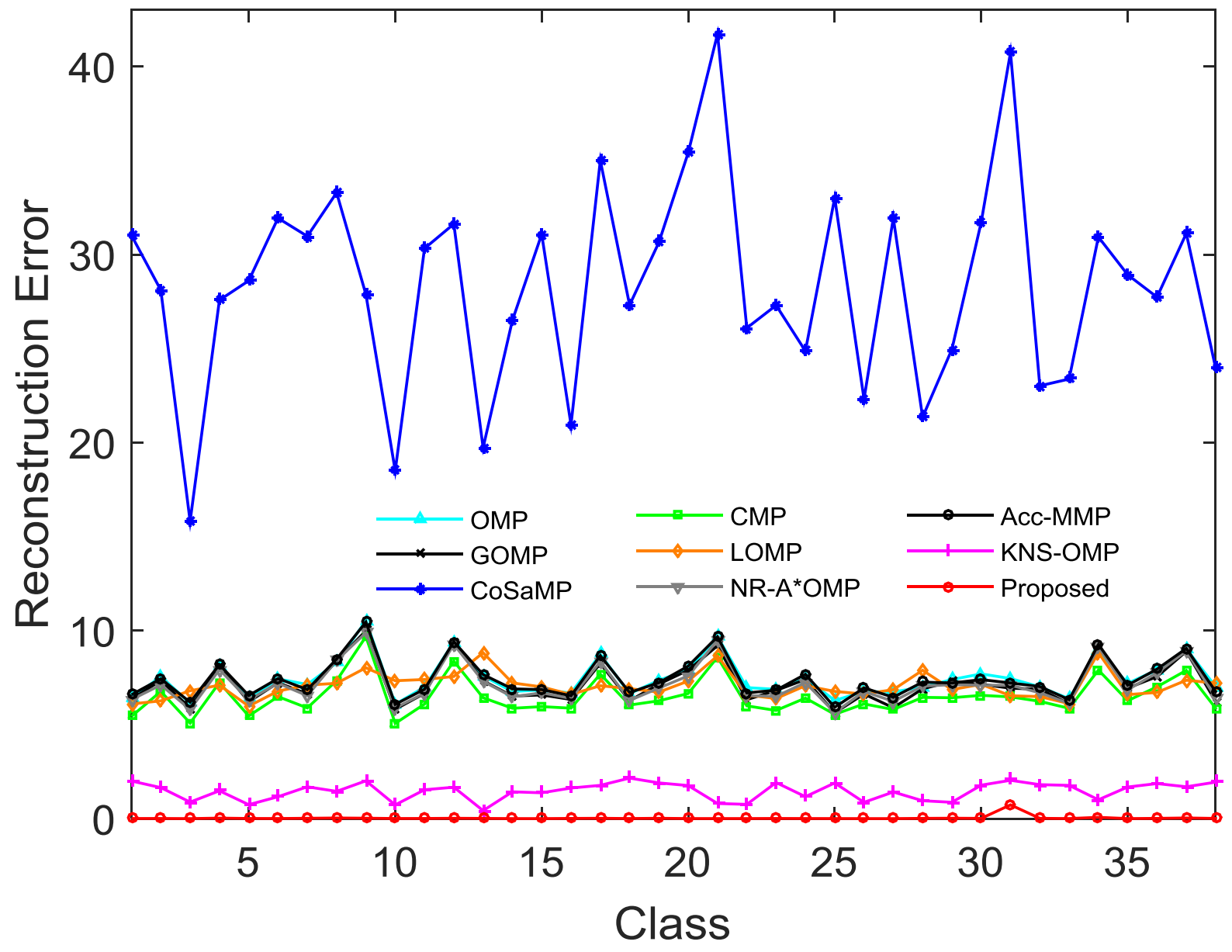

(b)

Fig. 4. Average reconstruction errors of images from each class using different matching pursuit algorithms on the extended Yale B face database. (a) results with randomly generated block occlusion. (b) results with real image block occlusion.

The advantages of the proposed method in tackling the occlusion issue are not only reflected in the competitive image reconstruction errors but also in the following two aspects: sparse coefficients and weight images. Here we take the 9th image of the first class with real image block occlusion as an example. The sparse coefficients of all the competing methods are shown in Fig. 5 from which we can see that the sparse coefficient of the proposed algorithm with $p = 1.7$ is the most sparse one. In this experiment, the more sparse the vector is, the less the method will be affected by occlusion, which further verifies that the proposed ITL-Correlation driven kernel non-second order minimization outperforms all the benchmarks.

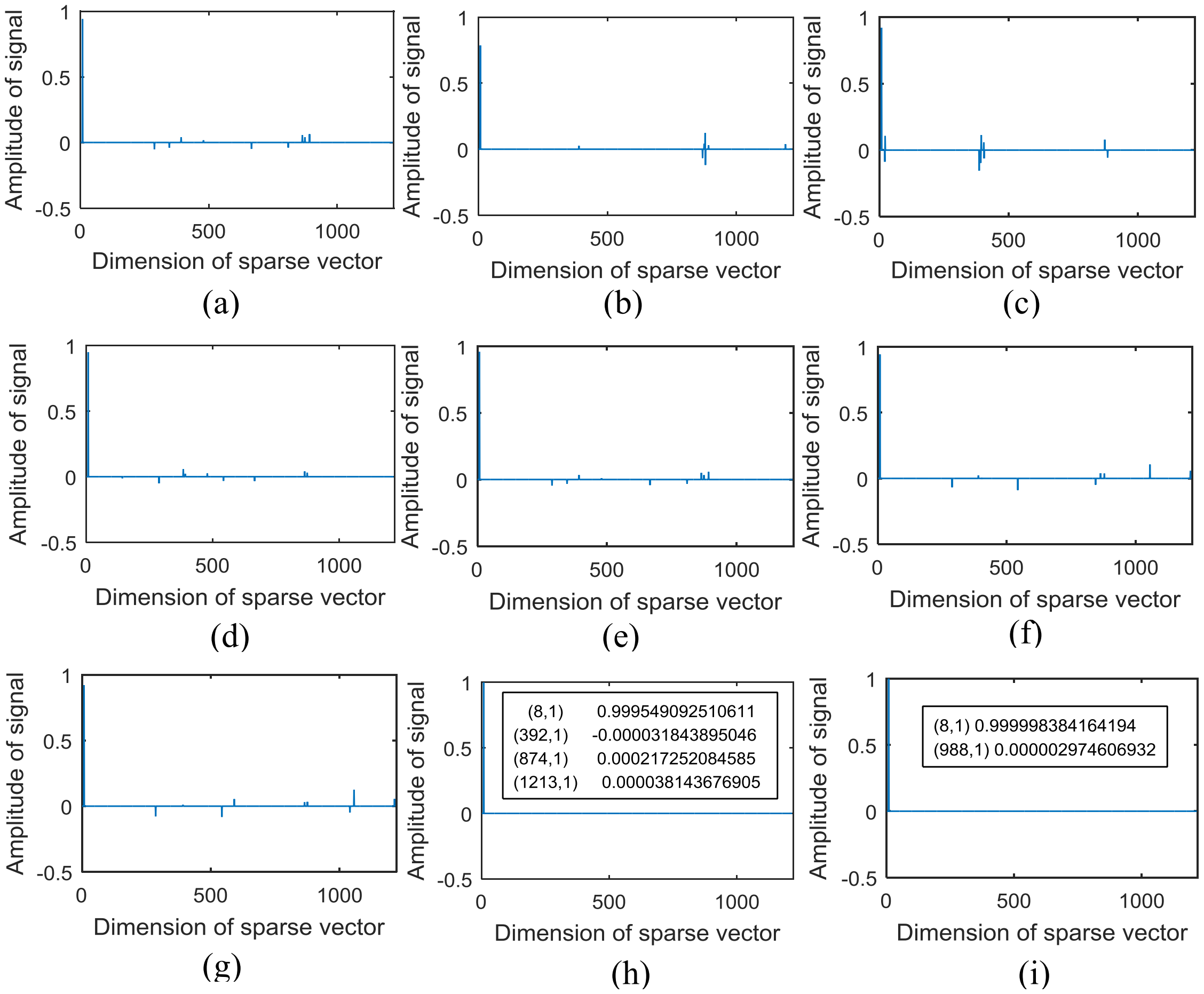

Fig. 5. Sparse vector of different matching pursuit algorithms. (a) OMP, (b) GOMP, (c) CoSaMP (d) CMP, (e) LOMP, (f) NR-A*OMP, (g) Acc-MMP, (h) KNS-OMP ($p$=1.7), (i) Proposed ($p$=1.7).

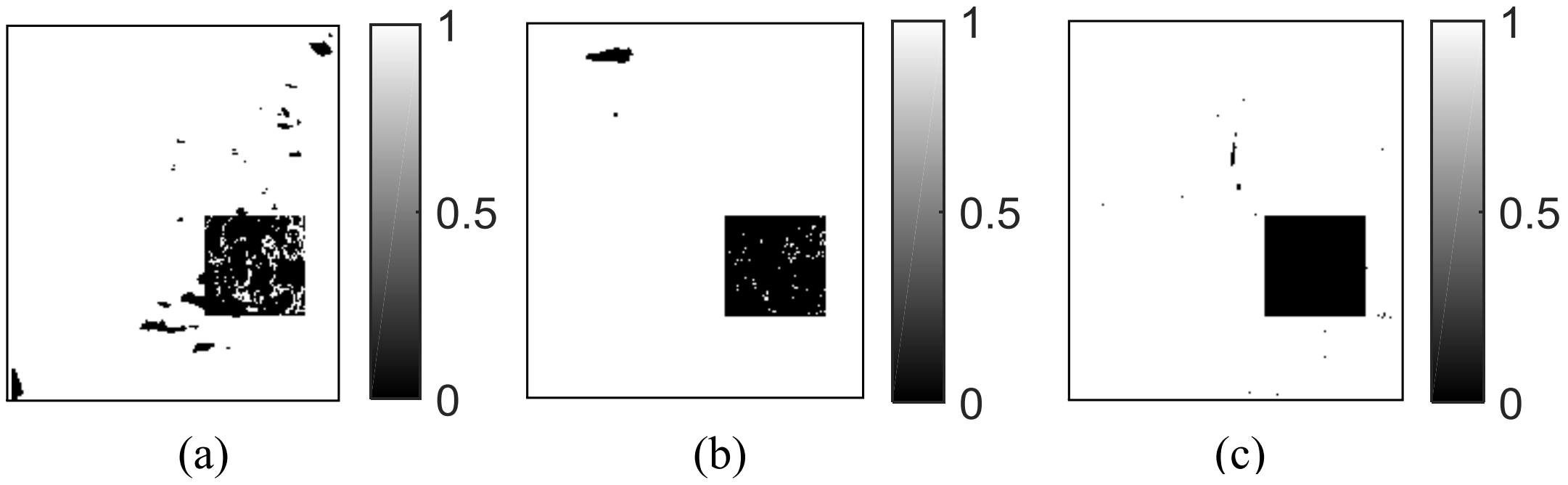

Fig. 6. Weight images from different ITL-based algorithms. (a) CMP, (b) KNS-OMP ($p$=1.7), (c) Proposed ($p$=1.7).

The occlusion resistance ability can also be verified from the weight images of the ITL-based algorithms in Fig. 6. To compare the weight images from different ITL-based algorithms clearly, we sorted the weight values of the three methods in ascending order and selected the first 10% of the smaller weight values and set them to zeros, and then set the remaining values to 1. To compare the weight learning ability of these algorithms, we need to find out which algorithm can accurately find the boundary between the main image content and the occlusion area. The weight image of the proposed method in Fig. 6 (c) has a smaller weight in the outlier area, which means that the proposed method can correctly detect and separate the outlier region and the normal image content by allocating small weights to the outlier part and large weights to the test image content. Although the algorithms in [24] and [27] can detect the occlusion region in Figs. 6 (a)-(b), there are still incorrectly-weighted pixels in the occlusion area, which means that the sparse vector learned from these algorithms will be perturbed by these incorrectly-weighted values.

## 4.3. Face Classification against Contiguous Occlusion

In this section, the proposed algorithm is used to verify the image classification ability on the images with contiguous occlusion. All the experiments are implemented on the extended Yale B face database with each image being resized to 96 × 84 pixels. Half of the images from each individual are selected (about 32 images) to construct dictionary $\boldsymbol{A}$, and the remaining images are used for testing. Therefore, the total numbers of images used for constructing the dictionary and for testing are 1210 and 1192, respectively.

The combination of sparse coding algorithms and a sparse representation classifier are usually used for classification [1, 21, 24]. Therefore, we consider the popular sparse representation classifier using the $l_1$ minimization method in [1] as the classifier for OMP, GOMP, CoSaMP, LOMP, NR-A*OMP, and Acc-MMP algorithms. For the CMP algorithm, we use the classifier used in [24] for comparison. For the KNS-OMP algorithm and the proposed algorithm, we use the classifier introduced in Algorithm 3 for classification. The same sparse level $L$ is utilized as the stopping criterion for all the competing algorithms for fair comparison. For all the ITL-based methods, we utilize the same stopping criterion during weight optimization: the norm of the current weight and the weight calculated from the previous iteration is less than $10^{-6}$.

First, we carry out experiment on the facial image with block occlusion to verify the classification performance of the proposed algorithm. To simulate the occlusion, we randomly replace a rectangular region in each testing image with an unrelated image. As used in [24], we use the baboon image for occlusion simulation in this experiment. Figs. 7(a)-(c) show the classification rates of testing image with 20% block occlusion. We carried out experiments on different downsampled images with sizes of $\{504, 896, 2016, 8064\}$ which correspond to the downsampling ratios of $\{1/4, 1/3, 1/2, 1\}$. Figs. 7(a)-(c) plot the classification rates when $L = 10, L = 20$, and $L = 30$, respectively. These three figures show that the proposed method obtains the best classification rate under any dimension of features, which means that the proposed algorithm is superior to other algorithms in suppressing the influence of occlusion.

To explore different order $p$ on the effect of classification performance, we plot curves for classification rates with changing feature size and sparse levels under different $p$ values in Figs. 7(d)-(f), where the classification rates with feature dimension 504 are the lowest ones, and those with feature dimension 8064 produce the best results. The results from Figs. 7(d)-(f) also give us important information that there exists a $p$ with $p < 2$ which gives better classification rates than that with $p = 2$. From the results in Figs. 7(d)-(f), we find that the best classification rates can be found from $p$ within $[1.5, 1.7]$.

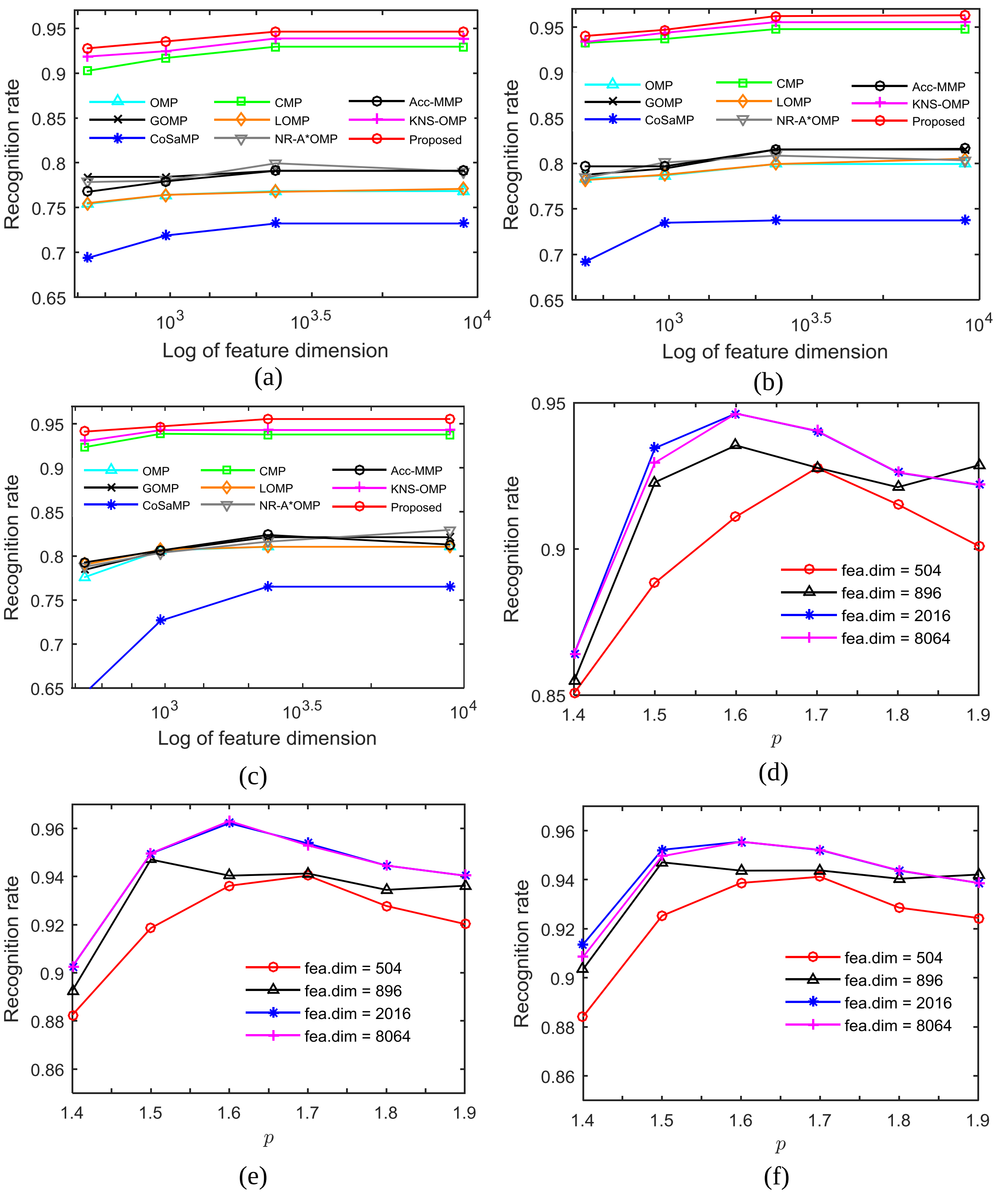

Fig. 7. Face classification rates under different feature dimensions and different $p$ values with 20% occlusion. (a)-(c) classification rates under different feature dimensions with $L = 10$, $L = 20$, and $L = 30$, respectively. (d)-(f) classification rates of the proposed method under different $p$ values with $L = 10$, $L = 20$, and $L = 30$.

Fig .8 plots how the classification rates vary according to the increase of the percentage of occlusion to evaluate the performance of the proposed algorithm. The facial images used for this experiment are resized to $48 \times 42$ pixels. The baboon image is resized to ensure that the test image has 10%, 20%, 30%, and 40% occlusions. Figs. 8(a)-(c) show the classification rates under $L = 10, L = 20$, and $L = 30$, respectively. We can see that

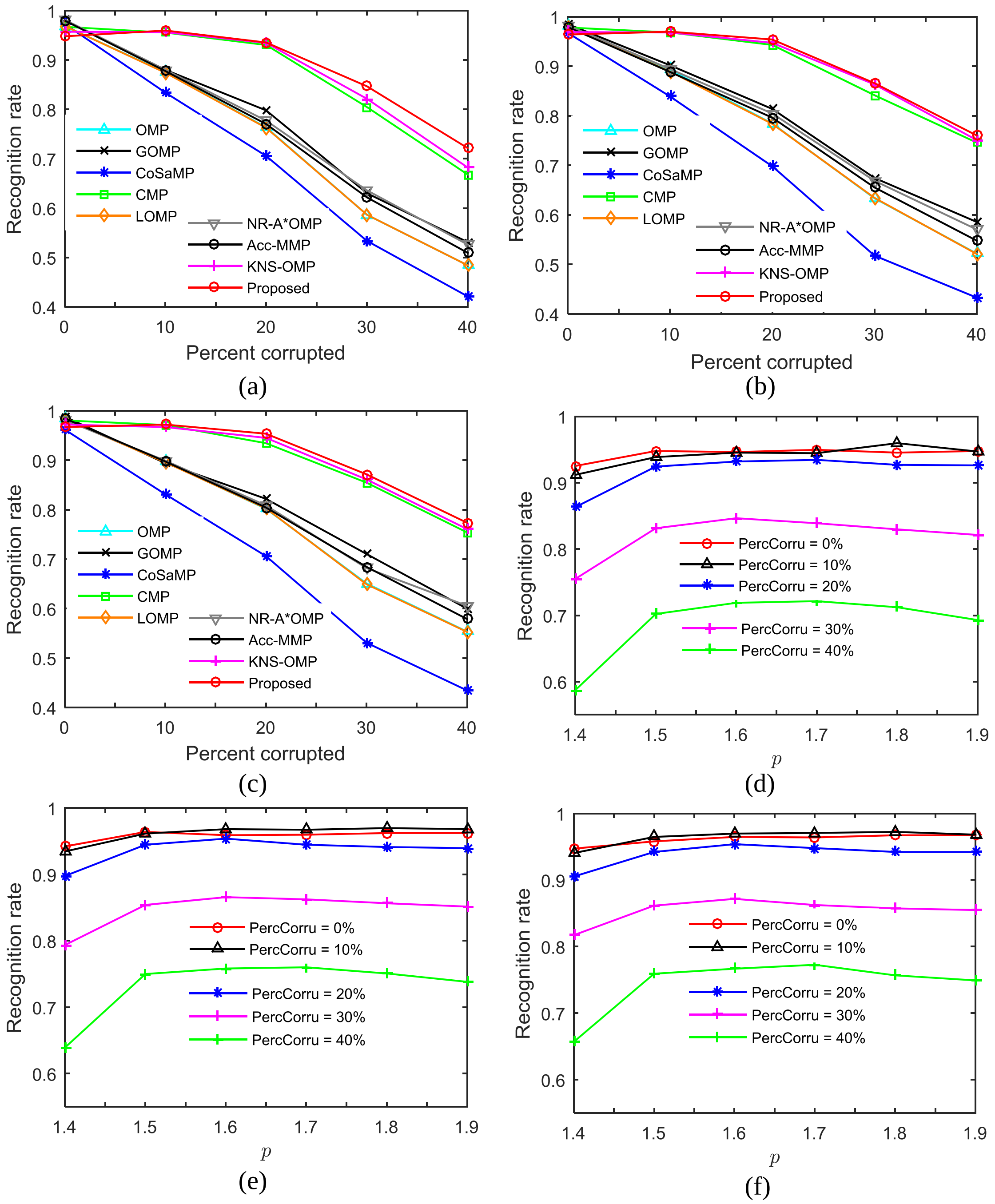

Fig.8. Face classification rates under different percentages of occlusion and different $p$ values. (a)-(c) comparison of classification rates of competing methods under different percentages of occlusion when $L = 10$, $L = 20$, and $L = 30$, respectively. (d)-(f) classification rates of the proposed method under different $p$ values with $L = 10$, L = 20, and $L = 30$.

the classification rates of OMP, GOMP, CoSaMP, LOMP, NR-A*OMP, and Acc-MMP decrease rapidly with the increase of the percentage of occlusions, while the rates of ITL-based methods decline gradually. Furthermore, with the increase on the level of occlusion, the outlier resistance ability of the proposed algorithm becomes increasingly apparent. This can be verified from Figs. 8(a)-(c) where the largest gap of the classification rate between the benchmarks and the proposed method increases from 13.25% to 32.19%. Figs. 8(d)-(f) plot the classification rates of the proposed algorithm with different percentages of occlusion under different $p$ values. The classification rates show that the proposed algorithm obtains the best results with $p$ within $[1.6, 1.7]$.

Table 4 reports the average classification rates and standard deviations of all the algorithms with different number of training samples over ten random trials. Test samples used in this experiment are resized to $48 \times 42$ with $20\%$ occlusion. The sparse level in this experiment is set to 20. From the results in Table 4, we know that the results of the proposed algorithm are the best ones under different numbers of training samples, and that increasing the number of training samples of each class will gradually widen the differences of the classification rates between the proposed algorithm and benchmarks.

Table 3. Classification rates of different algorithms on the extended Yale B database with different numbers of training samples (Nc). Ave. and Std. denote the average result and standard deviation, respectively. Best results are marked in bold.

| Method/Nc. | 20 | 24 | 28 | 32 |
|---|---|---|---|---|
| OMP | 0.7554±0.0091 | 0.7700±0.0104 | 0.7841±0.0067 | 0.7848±0.0026 |
| GOMP | 0.7798±0.0058 | 0.7999±0.0079 | 0.8218±0.0329 | 0.8137±0.0073 |
| CoSaMP | 0.7171±0.0318 | 0.7075±0.0073 | 0.7056±0.0069 | 0.7103±0.0058 |
| CMP | 0.9022±0.0028 | 0.9242±0.0035 | 0.9251±0.0045 | 0.9383±0.0022 |
| LOMP | 0.7557±0.0100 | 0.7703±0.0102 | 0.7838±0.0063 | 0.7848±0.0028 |
| NR-A*OMP | 0.7805±0.0062 | 0.7859±0.0087 | 0.7921±0.0062 | 0.7978±0.0065 |
| Acc-OMP | 0.7684±0.0066 | 0.7813±0.0046 | 0.7865±0.0043 | 0.7917±0.0062 |
| KNS-OMP | 0.9023±0.0024 | 0.9256±0.0056 | 0.9325±0.0019 | 0.9398±0.0031 |
| Proposed | **0.9064**±0.0024 | **0.9287**±0.0032 | **0.9397**±0.0034 | **0.9452**±0.0042 |

# 5. Conclusions

In this paper, we address the vulnerability problem of the existing OMP algorithms and investigate the relationship between the robustness of sparse representation and information theoretic learning (ITL) in the presence of outliers. With the analysis and discovery, we developed a novel OMP algorithm which incorporates the informatic-theoretic descriptors of entropy to the framework of OMP to change the internal feature identification and coefficient estimation mechanism, bringing robustness to learning coding coefficients. Since the proposed algorithm has inherent advantage for excluding the outlier information during optimization, it achieves superior performance in many realistic scenarios where there are noise and outliers in the data. We carried out experiments on both simulated data and real-world data with various noise and occlusion parameter settings. The experimental results consistently demonstrate that the proposed OMP algorithm is more robust than the state-of-the-art OMP algorithms in data recovery under noise of different distributions, and in image reconstruction and classification under different types of occlusions. Especially, the average reconstruction error has been reduced by 37% ~ 84% under different types of noise, and the classification accuracy has been increased by 20% ~ 24% with 20% occlusion in the data. The proposed algorithm has not taken the deep feature of the training data into consideration, thus a muti-layer OMP algorithm will be an interesting research in the future.

# Appendix

A: Proof:

$$V(\tau\boldsymbol{b},\boldsymbol{a}) = \frac{1}{\sqrt{2\pi}\sigma}\exp\left(-\frac{\|\tau\boldsymbol{b}-\beta\boldsymbol{a}\|_2^2}{2\sigma^2}\right)$$

$$= \frac{1}{\sqrt{2\pi}\sigma}\exp\left(|\tau|^2\frac{-\left\|\boldsymbol{b}-\frac{\beta}{\tau}\boldsymbol{a}\right\|_2^2}{2\sigma^2}\right)$$

$$= \frac{1}{\sqrt{2\pi}\sigma}\left(\exp\left(\frac{-\left\|\boldsymbol{b}-\frac{\beta}{\tau}\boldsymbol{a}\right\|_2^2}{2\sigma^2}\right)\right)^{|\tau|^2}$$

$$= \left(\frac{1}{\sqrt{2\pi}\sigma}\right)^{(1-|\tau|^2)} \left(\frac{1}{\sqrt{2\pi}\sigma} \exp\left(\frac{-\|\boldsymbol{b}-\beta_1^*\boldsymbol{a}\|_2^2}{2\sigma^2}\right)\right)^{|\tau|^2}$$

$$= \left(\frac{1}{\sqrt{2\pi}\sigma}\right)^{(1-|\tau|^2)} (V(\boldsymbol{b},\boldsymbol{a}))^{|\tau|^2}, \text{ where } \beta_1^* = \frac{\beta}{\tau}.$$

$$V(\boldsymbol{b},\tau\boldsymbol{a}) = \frac{1}{\sqrt{2\pi}\sigma} \exp\left(-\frac{\|\boldsymbol{b}-\beta\tau\boldsymbol{a}\|_2^2}{2\sigma^2}\right)$$

$$= \frac{1}{\sqrt{2\pi}\sigma} \exp\left(-\frac{\|\boldsymbol{b}-\beta_2^*\boldsymbol{a}\|_2^2}{2\sigma^2}\right)$$

$$= V(\boldsymbol{b},\boldsymbol{a}), \text{ where } \beta_2^* = \beta\tau.$$

B: Proof:

$$\|\boldsymbol{b}-\beta\boldsymbol{a}\|_2^2 = \|(\boldsymbol{b}-\beta^*\boldsymbol{a}) + (\beta^*\boldsymbol{a}-\beta\boldsymbol{a})\|_2^2$$

$$= \|\boldsymbol{b}-\beta^*\boldsymbol{a}\|_2^2 + \|\beta^*\boldsymbol{a}-\beta\boldsymbol{a}\|_2^2 + 2(\boldsymbol{b}-\beta^*\boldsymbol{a})^T(\beta^*\boldsymbol{a}-\beta\boldsymbol{a}),$$

With some linear algebraic operation and considering $\beta^* = \boldsymbol{a}^T\boldsymbol{b}(\boldsymbol{a}^T\boldsymbol{a})^{-1}$ and $\boldsymbol{a}^T\boldsymbol{b} - (\boldsymbol{a}^T\boldsymbol{a})\beta^* = \boldsymbol{0}$, the third term above can be written as:

$$(\boldsymbol{b}-\beta^*\boldsymbol{a})^T(\beta^*\boldsymbol{a}-\beta\boldsymbol{a}) = (\boldsymbol{b}-\beta^*\boldsymbol{a})^T\boldsymbol{a}(\beta^*-\beta) = \boldsymbol{0}$$

Thus, we have

$$\|\boldsymbol{b}-\beta\boldsymbol{a}\|_2^2 = \|\boldsymbol{b}-\beta^*\boldsymbol{a}\|_2^2 + \|\beta^*\boldsymbol{a}-\beta\boldsymbol{a}\|_2^2$$

$$= \|\boldsymbol{b}-\beta^*\boldsymbol{a}\|_2^2 + \|(\beta^*-\beta)\boldsymbol{a}\|_2^2.$$

Since the term $\|(\beta^*-\beta)\boldsymbol{a}\|_2^2$ is non-negative, we have

$$\|\boldsymbol{b}-\beta\boldsymbol{a}\|_2^2 \geq \|\boldsymbol{b}-\beta^*\boldsymbol{a}\|_2^2,$$

which shows that there exists a $\beta^*$ that minimizes $\|\boldsymbol{b}-\beta\boldsymbol{a}\|_2^2$. If the equality $\|\boldsymbol{b}-\beta\boldsymbol{a}\|_2^2 = \|\boldsymbol{b}-\beta^*\boldsymbol{a}\|_2^2$ holds, then we have $\|(\beta^*-\beta)\boldsymbol{a}\|_2^2 = \boldsymbol{0}$, which means $\beta^* - \beta = 0$, i.e., $\beta = \beta^*$, since $\boldsymbol{a}$ is a non-zero vector. Thus, the only $\beta$ that makes the equality $\|\boldsymbol{b}-\beta\boldsymbol{a}\|_2^2 = \|\boldsymbol{b}-\beta^*\boldsymbol{a}\|_2^2$ hold is $\beta = \beta^*$. Then we can conclude that vector $\boldsymbol{b}$ can be represented by $\boldsymbol{a}$ under a unique coefficient $\beta$.

# Acknowledgements

This work is supported in part by Australian Research Council (ARC) under Discovery Grants DP140101075.